%% file: main.tex
\documentclass{article}
\usepackage[utf8]{inputenc}
\usepackage[margin=8em]{geometry}
\usepackage{graphicx}
\usepackage{amsmath}
\usepackage{amsthm}
\usepackage{enumitem}
\usepackage{booktabs}
\usepackage{makecell}
\usepackage{hyperref}
\usepackage{cleveref}
\usepackage{mathtools}

\usepackage{tikz}
\usetikzlibrary{decorations.pathreplacing,shapes.misc,bending}
\usepackage{tkz-euclide}
\usepackage{environ}
\makeatletter
\newsavebox{\measure@tikzpicture}
\NewEnviron{scaletikzpicturetowidth}[1]{%
  \def\tikz@width{#1}%
  \def\tikzscale{1}\begin{lrbox}{\measure@tikzpicture}%
  \BODY
  \end{lrbox}%
  \pgfmathparse{#1/\wd\measure@tikzpicture}%
  \edef\tikzscale{\pgfmathresult}%
  \BODY
}
\makeatother

\newtheoremstyle{exampstyle}
  {0.2em} 
  {0.em} 
  {} 
  {} 
  {\bfseries} 
  {.} 
  {.5em} 
  {} 
\theoremstyle{exampstyle} \newtheorem{thm}{Theorem}
\theoremstyle{exampstyle} 

\usepackage{authblk}
\usepackage{natbib}

\input{preamble}

\usepackage{url}
\title{The Effects of Regularization and Data Augmentation are\\Class Dependent}
\author{Randall Balestriero$^1$}
\author{L\'eon Bottou$^1$}
\author{Yann LeCun$^{1,2}$}
\affil{$^1$Meta~AI~Research, $^2$NYU\\{\footnotesize \{rbalestriero,leonb,ylecun\}@fb.com}}
\date{}

\begin{document}

\maketitle

\begin{center}
    Codes, full result tables, and pre-trained model weights will be released soon
\end{center}

\begin{abstract}
    Regularization is a fundamental technique to prevent over-fitting and to improve generalization performances by constraining a model's complexity. Current Deep Networks heavily rely on regularizers such as 
    Data-Augmentation (DA) or weight-decay, and employ structural risk minimization, i.e. cross-validation, to select the optimal regularization hyper-parameters. In this study, we demonstrate that techniques such as DA or weight decay produce a model with a reduced complexity that is unfair across classes. The optimal amount of DA or weight decay found from cross-validation leads to disastrous model performances on some classes e.g. on Imagenet with a resnet50, the ``barn spider'' classification test accuracy falls from $68\%$ to $46\%$ only by introducing random crop DA during training. Even more surprising, such performance drop also appears when introducing uninformative regularization techniques such as weight decay. Those results demonstrate that our search for ever increasing generalization performance -averaged over all classes and samples- has left us with models and regularizers that silently sacrifice performances on some classes. This scenario can become dangerous when deploying a model on downstream tasks e.g. an Imagenet pre-trained resnet50 deployed on INaturalist sees its performances fall from $70\%$ to $30\%$ on class \#8889 when introducing random crop DA during the Imagenet pre-training phase. Those results demonstrate that designing novel regularizers without class-dependent bias remains an open research question.
\end{abstract}

\section{Introduction}

Machine learning and deep learning aim at learning systems to solve as accurately as possible a given task at hand \citep{lecun1998gradient,bishop2006pattern,jordan2015machine}. This process often takes the form of being given a {\em finite training set} and a {\em performance measure}, optimizing the system's parameters e.g. from gradient updates, and assessing the system's performance on test set samples, i.e. samples that were not used during the system optimization. {\bf As the training set is finite, and the optimal design of the system is unknown, it is common to employ regularization during the optimization phase to reduce over-fitting} \citep{tikhonov1943stability,tihonov1963solution} i.e. to decrease the system's performance gap between train set and test set samples \citep{simard1991tangent,chapelle2000vicinal,bottou2012stochastic,neyshabur2014search}.

Data-Augmentation (DA) is a data-driven and informed regularization strategy that artificially increase the number of training samples \citep{shorten2019survey}. As opposed to most {\em explicit} regularizers e.g. Tikhonov regularization \citep{krogh1991simple}, also denoted as weight decay, DA's regularization is {\em implicit} as it is not a function of a model's parameter, but a function of the training samples \citep{neyshabur2014search,hernandez2018data,lejeune2019implicit}; although some DA strategies can be turned into explicit regularizers \citet{balestriero2022data}. Nevertheless, a key distinction between DA and weight decay is that DA requires more domain knowledge to be successful than weight decay. Most ---if not all--- of current state-of-the-art employ such regularizers \citep{huang2018condensenet,chen2020improved,liu2021swin,tan2021efficientnetv2,liu2022convnet}.

In this paper, we will demonstrate that {\bf when employing regularization such as DA or weight decay, a significant bias is introduced into the trained model}. In particular, {\bf the regularized model exhibits strong per-class favoritism i.e. while the average test accuracy over all classes improves when employing regularization, it is at the cost of the model becoming arbitrarily inaccurate on some specific classes} as illustrated in \cref{fig:teaser}. After a brief theoretical justification on why and when DA can be the cause of bias (\cref{sec:theory}), we propose a dedicated sensitivity analysis of the bias produced by different amounts of DA in \cref{sec:pretrained,sec:quantify}, which is followed by a similar study dedicated to weight decay in \cref{sec:decay} and transfer learning in \cref{sec:transfer}. \textit{We shall highlight that although we perform a class-level study, it is possible to refine this entire analysis at the sample-level.}

\input{tikz_figure}

For readers familiar with statistical estimation results e.g. the bias-variance trade-off \citep{kohavi1996bias,von2011statistical} or bayesian estimation e.g. Tikhonov regularization \citep{box2011bayesian,gruber2017improving}, it should not be surprising that regularization produces bias. In fact, it is often beneficial to introduce bias through regularization if it results in a significant reduction of the estimator variance ---when one aims to minimize the average empirical risk. This is one of the main reason behind the success of techniques such as ridge regression. However, what is potentially dangerous is that {\bf the bias introduced by regularization treats classes differently, including on transfer learning tasks} as we will demonstrate in \cref{sec:transfer}.
Those observations also support recent theoretical results tying a model's performance to its robustness and to DA, as we discuss in \cref{sec:more_theory}.

\section{Regularization Creates Class-Dependent Model Bias that can be Harmful even for Transfer Learning Tasks}
\label{sec:DA}

The first part of our study focuses on DA, a technique that regularizes a model by introducing new training samples, derived from the observed ones. DA samples have been known to sometimes disregard the semantic information of the original samples \citep{krizhevsky2012imagenet}. Nevertheless, DA remains applied universally, and fearlessly across tasks and datasets \citep{shorten2019survey} as it provides significant performance improvements, even in semi-supervised and unsupervised settings \cite{guo2018deep,xie2020unsupervised,misra2020self}. We first provide in \cref{sec:theory,sec:pretrained} some intuition on why DA can be a source of bias regardless of the task, dataset and model at hand. We then quantify the amount of bias caused by DA in various realistic scenarios in \cref{,sec:quantify}; and extend our analysis to weight decay in \cref{sec:decay}. Finally, we conclude by demonstrating how the bias introduced by regularization transfers to downstream tasks e.g. when deploying an Imagenet ({\em source}) trained model on the INaturalist ({\em target}) dataset in \cref{sec:transfer}; that scenario is key as it demonstrates the potential harm of selecting the best performing model ---on average--- on the source dataset which could turn out to also be the most biased model on the target dataset class of interest. \textit{ In fact, it is crucial to remember that regularization, or any other form of structural risk minimization, improves generalization performances by increasing the bias of the estimator so that the estimator's variance is decreased by a greater amount. However, nothing guarantees the fairness of this bias i.e. for it to be equally distributed amongst the dataset classes.}

\subsection{When Data-Augmentation Creates Bias}
\label{sec:theory}

To provide a simple explanation on how DA causes bias in a trained model, we propose the following derivation that holds for any signal e.g. timeseries, images, videos. Without loss of generality \citep{hui2020evaluation} we will consider here the $\ell_2$ loss, although the same derivation carries out with any desired metric. 

{\bf Dataset notations.}~Given a sample $\vx \in \mathcal{X}$ with $\mathcal{X} \subset \mathbb{R}^{D}$, we consider $\vy \triangleq f^*(\vx)$ to be the ground-truth target value. Hence our hope is to learn an approximator $f_{\theta}$ that is as close as possible to $f^*$ everywhere in $\mathcal{X}$, although we only observe a finite training set $\sX\triangleq \{(\vx_1,\vy_1),\dots,(\vx_N,\vy_N)\}$.\\
{\bf Data-Augmentation notations.}~Additionally, one employs a DA policy $\mathcal{T}:\mathbb{R}^{D}\times \mathcal{K}\mapsto \mathbb{R}^{D}$ such that given a transformation parameter $\alpha \in \mathcal{K}$, $\mathcal{T}_{\alpha}(\vx)$ produces the transformed view of $\vx$. Often, one also defines a density $p$ on $\mathcal{K}$ that helps in sampling transformation parameters that are a priori known to be the most useful.

\begin{thm}
\label{thm:bias}
Whenever the transformations produced by $\mathcal{T}_{\alpha},\forall \alpha$ do not respect the level-set of $f^*$, and whenever the model has enough capacity to minimize the training loss, the DA will create irreducible bias in $f_{\theta}$ as in
\begin{align}
    \underbrace{\sum_{(\vx,\vy) \in \sX}\mathbb{E}_{\alpha}\left[ \|\vy -f^*(\mathcal{T}_{\alpha}(\vx))\|_2^2 \right]}_{\mathclap{\text{\hspace{0.5cm}$=0$ iff the DAs of $\vx$ are on the same level-set of $f^*$}}} > 0 \text{ and } \underbrace{\sum_{(\vx,\vy) \in \sX}\mathbb{E}_{\alpha}\left[\|\vy-f_{\theta}(\mathcal{T}_{\alpha}(\vx))\|_2^2\right]=0}_{\text{zero training error}}\implies \text{ biased  } f_{\theta}.\label{eq:ratio}
\end{align}
\end{thm}

The main idea of the proof, provided in \cref{sec:proof_bias}, is to show that if a transformation does not move samples on the level-set of the true function (left-hand-side of \cref{eq:ratio}), then $f_{\theta}$ will learn a different level (since it has $0$ training error), and thus $\|f^*-f_{\theta}\|>0$ i.e. $f_{\theta}$ is biased regardless of the training set.

Whenever the left-hand-side of \cref{eq:ratio} is $0$, the DA is denoted as {\em label-preserving} \citep{cui2015data,taylor2018improving}. From the above, we see that {\bf unless the target $\vy$ associated to $\mathcal{T}_{\alpha}(\vx)$ is modified accordingly to encode the shift in the target function level-set produced by $\mathcal{T}_{\alpha}$, any DA that is not label-preserving will introduce a bias}. Some DAs propose to incorporate label transformation i.e. not only $\vx$ but also $\vy$ is augmented to better inform on the uncertainty that has been added into $\mathcal{T}_{\theta}(\vx)$. This is for example the case for MixUp \citep{zhang2017mixup}, ManifoldMixUp \citep{verma2019manifold}, CutMix \citep{yun2019cutmix} and their extensions.

\begin{figure}[t!]
    \centering
    \includegraphics[width=\linewidth]{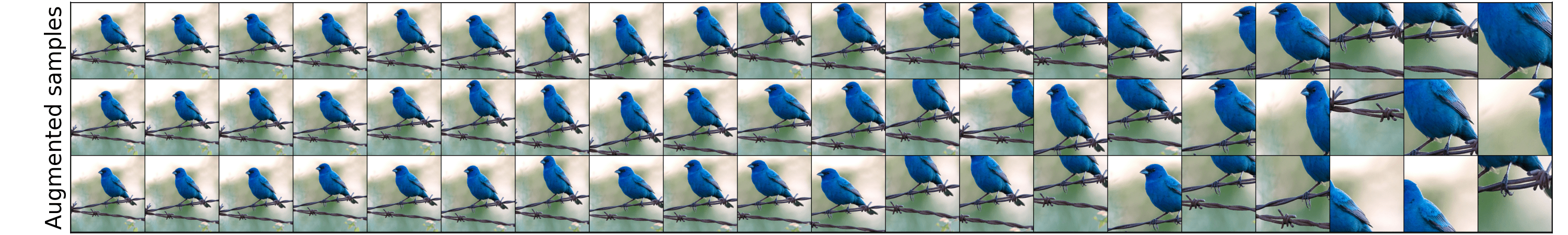}\\[0em]
    \includegraphics[width=\linewidth]{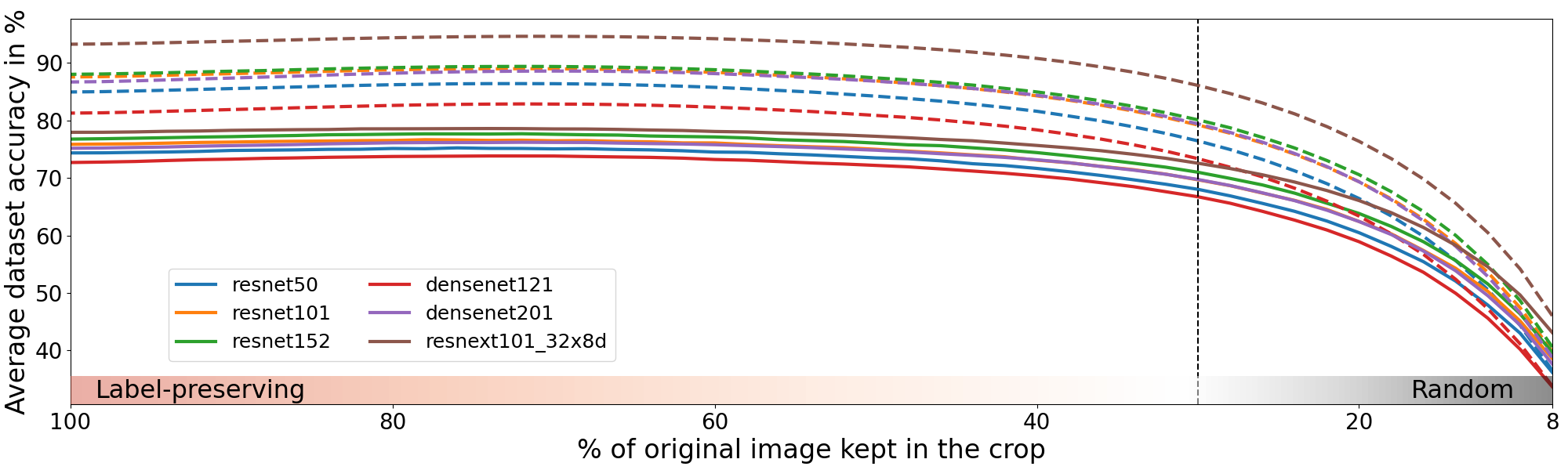}\\
    \includegraphics[width=0.66\linewidth]{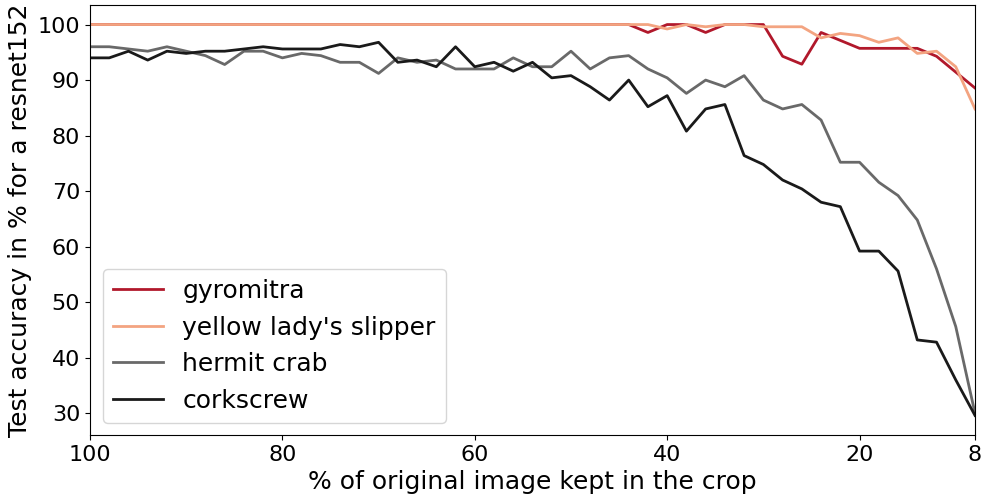}
    \includegraphics[width=0.33\linewidth]{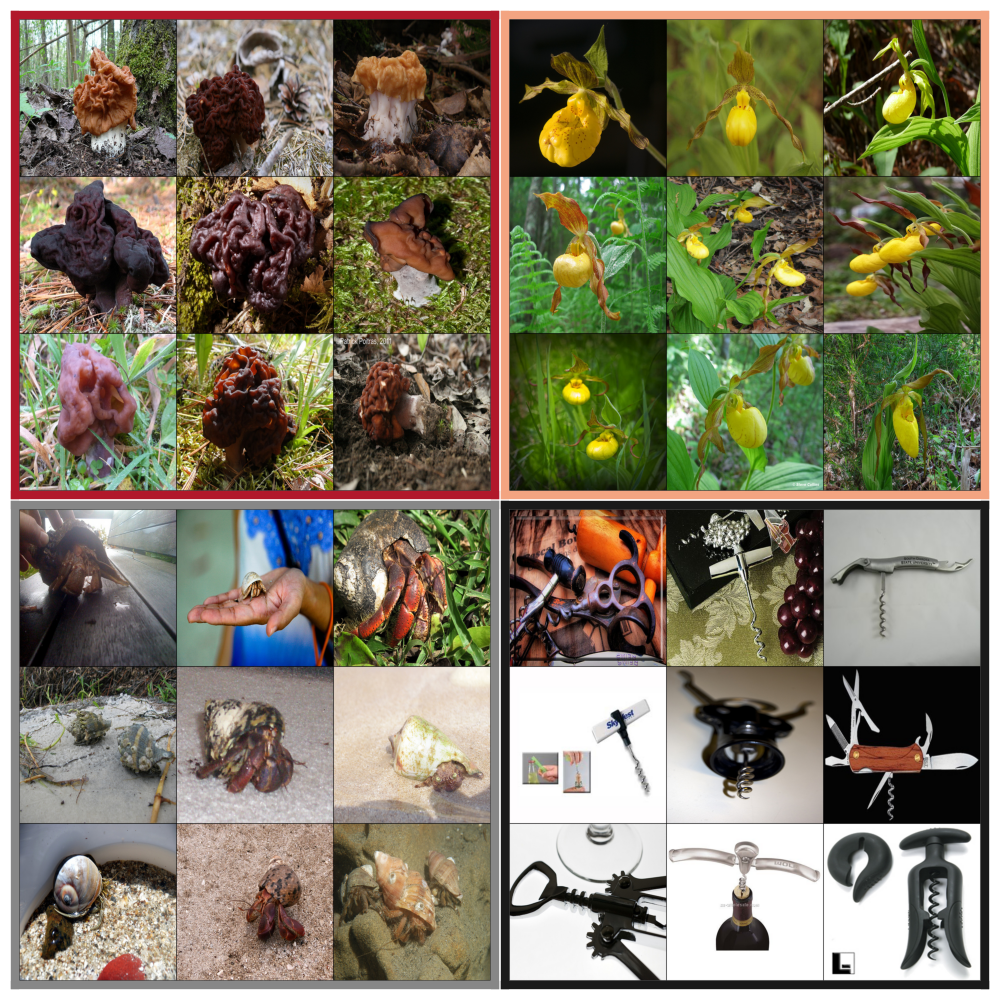}
    \vspace{-0.7cm}
    \caption{\small \textit{All results in this figure employ official \underline{pretrained} models from PyTorch with random crop DA.} We present examples of an augmented image of class ``bird'' (\textbf{top}) along with the average accuracy on the training set (dashed line) and test set (plain line) on Imagenet, using $6$ popular architectures (\textbf{middle}). The random crop DA seems to loose its label-preserving property when less than 30\% of the image is kept in the crop. However, when looking at per-class performances we observe an entirely different story where {\bf random crop DA can be label-preserving with only 8\% of the original image for some classes, while for other classes the label information starts to reduce at around 50\%} as reported at the \text{bottom} along with $9$ images of the corresponding classes. The CutOut and ColorJitter cases are presented in \cref{fig:dataset_erasing,fig:dataset_color} and exhibit the same per-class behaviors.}
    \label{fig:dataset_accuracy}
\end{figure}

Our goal in the next \cref{sec:pretrained} is to demonstrate how DAs such as random crop, color jittering, or CutOut are only label preserving for some values of $\alpha$ that vary with the sample class. As a consequence, while the use of the DA improves the average test performance, it is at the cost of a significant reduction in performance for some of the classes.

\subsection{The Same Data-Augmentation can be Label-Preserving or Not Between Different Classes}
\label{sec:pretrained}

In the previous \cref{sec:theory} we provided a general argument on the sufficient conditions for DA to produce a biased model. We hope in this section to provide a more concrete example that applies to current DN training. To that end, we will demonstrate that a DA can be label-preserving or not depending on the sample's class, hence, since the same DA policy is employed for all classes, the augmented dataset will exhibit a class-imbalance in favor of the classes for which the DA is most label-preserving.

To measure by how much a given DA, $\mathcal{T}_{\alpha}$, is label-preserving, we propose to take $6$ popular architectures that are pre-trained on Imagenet \citep{deng2009imagenet} from the official PyTorch \citep{paszke2019pytorch} repository, and to evaluate their accuracy performances for varying DA settings (top of \cref{fig:dataset_accuracy}). We observe that when considering the dataset as a whole, it is possible to identify a DA regime as a function of $\alpha$ for which the amount of information present in $\mathcal{T}_{\alpha}(\vx)$ becomes insufficient to predict the correct label on average. But more interestingly, we also take the per-class accuracy performance (bottom of \cref{fig:dataset_accuracy}) and observe that for some classes, any level of transformation $\alpha$ can produce augmented samples with enough information to be correctly classified, while other classes see their samples become unpredictable as soon as $\mathcal{T}_{\alpha}$ moves away from the identity mapping. Note that we report test set performances ensuring that the observed performances are not due to ad-hoc memorization. 

\begin{figure}[t!]
    \centering
    \begin{minipage}{0.32\linewidth}
    \centering
    \underline{random crop}
    \includegraphics[width=\linewidth]{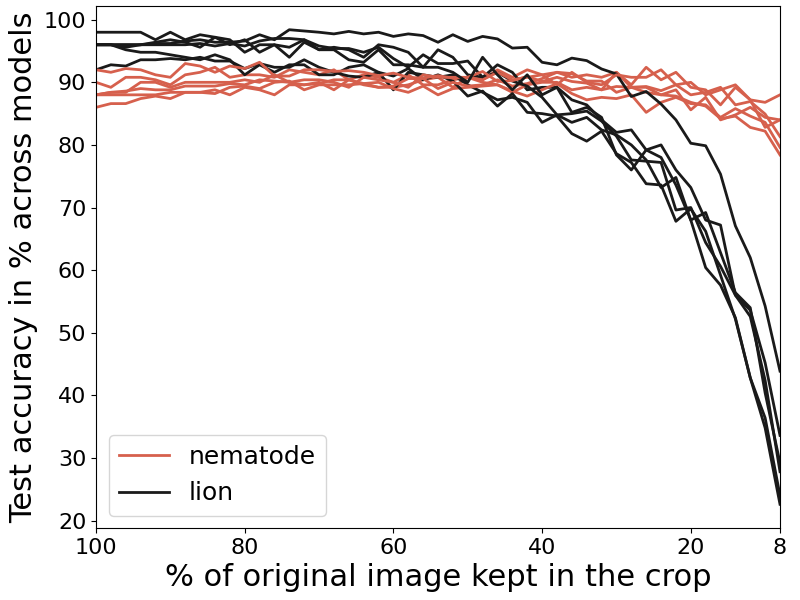}
    \end{minipage}
    \begin{minipage}{0.32\linewidth}
    \centering
    \underline{CutOut}
    \includegraphics[width=\linewidth]{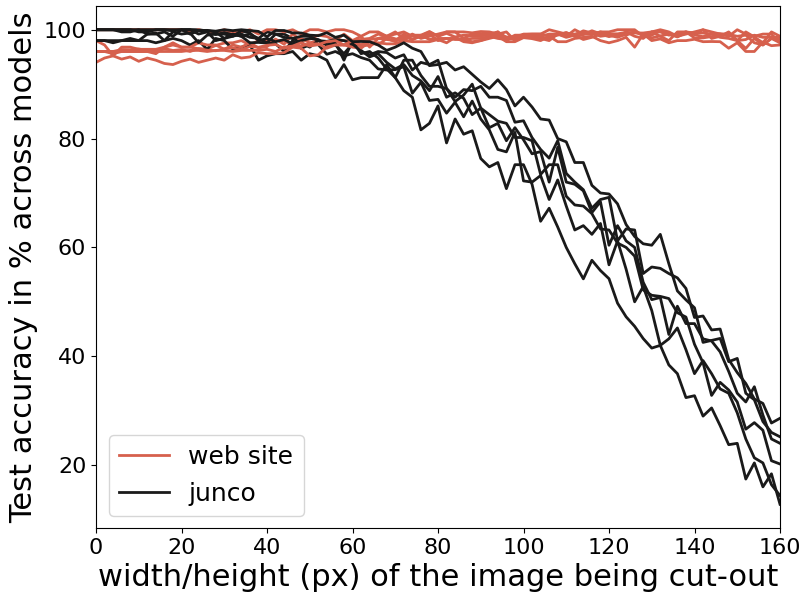}
    \end{minipage}
    \begin{minipage}{0.32\linewidth}
    \centering
    \underline{color jittering}
    \includegraphics[width=\linewidth]{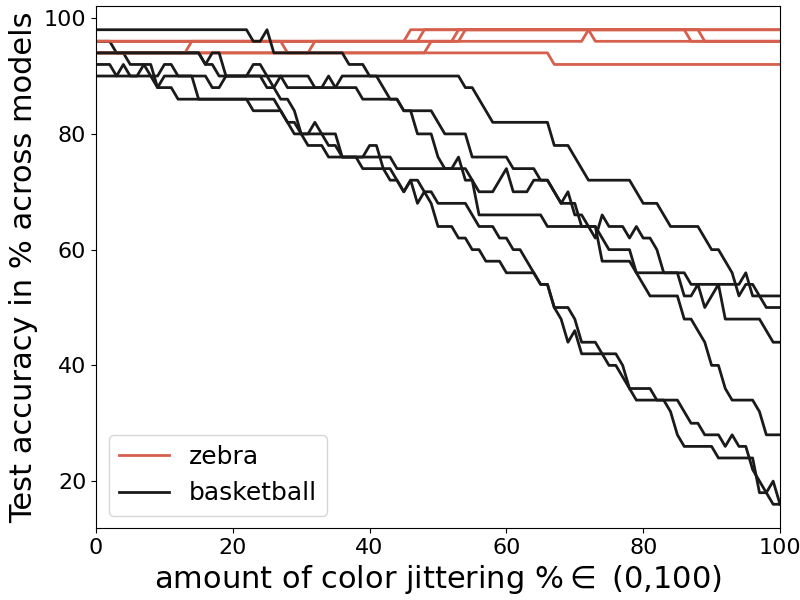}
    \end{minipage}
    \vspace{-0.2cm}
    \caption{\small \textit{All results in this figure employ official \underline{pretrained} models from PyTorch.} Reprise of the bottom left of \cref{fig:dataset_accuracy} for three different DAs (\textbf{each column}) and using the same $6$ popular architectures (resnet50,
        resnet101,
        resnet152,
        densenet121,
        densenet201,
        resnext101-32x8d) (\textbf{different lines}). We observe that {\bf across DAs, different architectures agree on the label-preserving regimes for $\mathcal{T}_{\alpha}$ i.e. even an ensemble of model would not reduce the class-dependent bias of the final prediction}.}
    \label{fig:allmodels}
\end{figure}

To further ensure that the observed relation between label-preservation, sample class, and amount of transformation $\alpha$ is sound, we provide in \cref{fig:allmodels} the per-class test accuracy on different models, all exhibit the same trends. In short, we identify that \textbf{when creating an augmented dataset by applying the same DA across classes, the number of per-class samples that actually contain enough information about their true labels will become largely imbalance between classes, even if the original dataset was balanced}. Any model trained on the augmented dataset will thus focus on the classes for which the DA is the most label-preserving.
We propose in the next section to precisely quantify the impact of DA on each dataset class.

\subsection{Measuring the Average Treatment Effect of Data-Augmentation on Models' Class-Dependent Bias}
\label{sec:quantify}

This section aims at quantifying precisely the amount of downward or upward per-class performance shift that came as a result from using DA. We thus propose a sensitivity analysis by training a large collection of models with varying DA policies to precisely assess the relation between DA and class-dependent model bias.

\begin{figure}[t!]
    \centering
    \includegraphics[width=\linewidth]{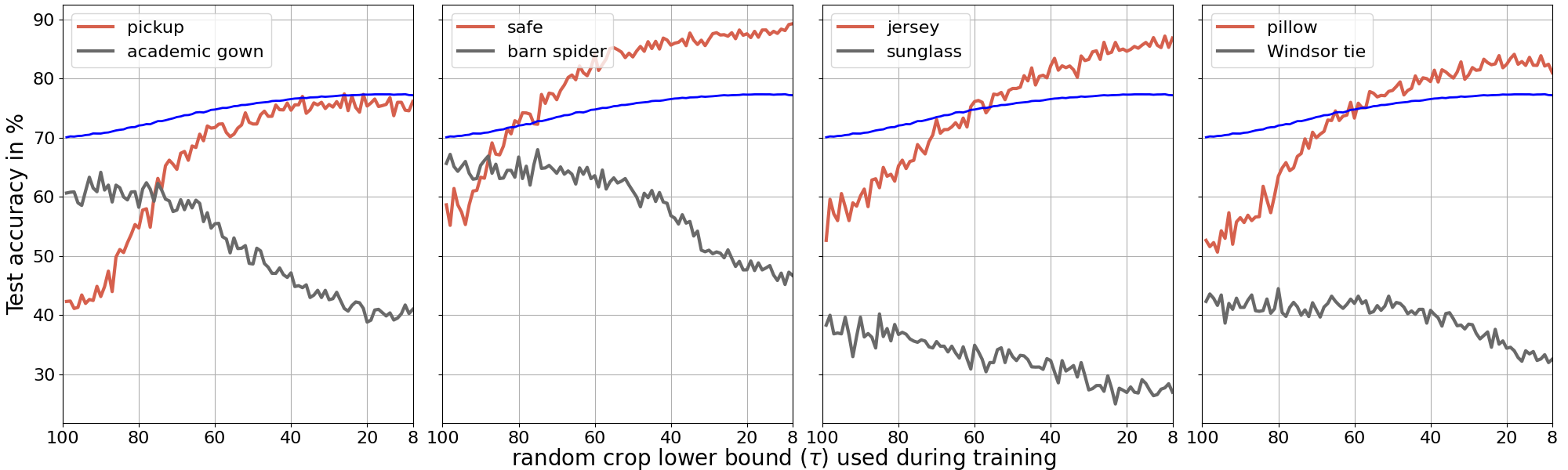}
    \vspace{-0.7cm}
    \caption{\small \textit{All results in this figure are \underline{ averaged over 20 runs} and employ the official \underline{resnet50} implementation from PyTorch that we trained on Imagenet with horizontal flip and varying lower bound for the \underline{random crop} DA.} During training, the lower bound on the amount of the original image kept in the random crops (x-axis) varies from 100\% to 8\% (commonly employed value), the upper bound is always 100\%, and the test images are obtained from center crops.  We observe that \textbf{training the same architecture but with varying random crop parameter (lower bound) provides greater average test accuracy (blue) but makes the per-class performance fall for some of the classes}. Samples for each class are provided in \cref{fig:resnet50_crop_samples}, in the appendix.}
    \label{fig:bias}
\end{figure}

First, we propose in \cref{fig:bias} a sensitivity analysis by training the same architecture on Imagenet with varying DA policies. In particular, we consider a given DA (random crop in this case) and we vary the support of the parameter $\alpha$ which represents how much of the original image is kept in the crop. We train DNs using $\alpha \in [100,\tau]$ with $\tau$ varying from $100$ to $8$ and for each case, we report our metrics averaged over $20$ trained models. We observe a clear relation between {\bf increase in the strength of the DA, increase in the average test accuracy overall classes, and decrease in some per-class test accuracies}. For example, on a resnet50 Imagenet setting, the accuracy on the ``academic gown'' class goes from 62\% to 40\% steadily as $\tau$ decreases. We defer the same experiment but using weight decay in the next \cref{sec:decay}.

To further convey our claim, we now propose a formal statistical test \citep{neyman1933ix,fisher1955statistical} on the hypothesis that the per-class accuracy is significantly lower when DA is applied for those classes. A test of significance is a formal procedure for comparing observed data with a claim or hypothesis. In our case we aim to test if the mean accuracy on class $y$ of a model trained without DA is greater than the one obtained with DA. Due to the stochastic optimization process, this accuracy is a random variable even when the dataset is not changed. Hence we define that random variable by ${\rm Accu}_y(\sX)$ and our null hypothesis by $H_0=\mathbb{E}[{\rm Accu}_y(\sX)]<\mathbb{E}[{\rm Accu}_y(\sX_{\rm DA})]$ with $\sX$ the original dataset and $\sX_{\rm DA}$ the DA dataset using random crop with $\alpha \in [100,8]$.
A one-sided t-test is used and the form that does not assume equal variances is known as Welch’s t-test \citep{welch1947generalization} with statistic
\begin{align*}
    t=\frac{\hat{\mu}(\sX_1)-\hat{\mu}(\sX_2)}{\sqrt{\hat{\sigma}^2(\sX_1)-\hat{\sigma}^2(\sX_2)}},
\end{align*}
and with degrees of freedom
$
    \nu=\left(\frac{\hat{\sigma}^2(\sX_1)}{N_1}+\frac{\hat{\sigma}^2(\sX_2)}{N_2}\right)^2\left(\frac{\hat{\sigma}^4(\sX_1)}{N_1^2(N_2-1)}+\frac{\hat{\sigma}^4(\sX_2)}{N_2^2(N_1-1)}\right)^{-1}
$.
We obtain that there is enough evidence to reject $H_0$ with $95\%$ confidence for $4.4\%$ of all the classes, and with $99\%$ confidence for $2.1\%$ of all the classes. Hence there is sufficient evidence to say that the per-class test accuracies is not increased when introducing DA for $4.4\%$ of the $1000$ Imagenet classes.
Note that one could formulate this problem in term of average treatment effect, where the treatment is the application of DA and the outcome is the accuracy on class $y$ of the trained model. Doing so, one could measure the per-sample bias from the Conditional Treatment Effect, however we limit ourselves to a per-class study and leave such fine-grained analysis for future work. The next \cref{sec:decay} proposes to reproduce those experiments but considering weight decay as a regularizer instead of DA.

\subsection{Uninformed Weight Decay Also Creates Class-Dependent Model Bias}
\label{sec:decay}

\begin{figure}[t!]
    \centering
    \includegraphics[width=\linewidth]{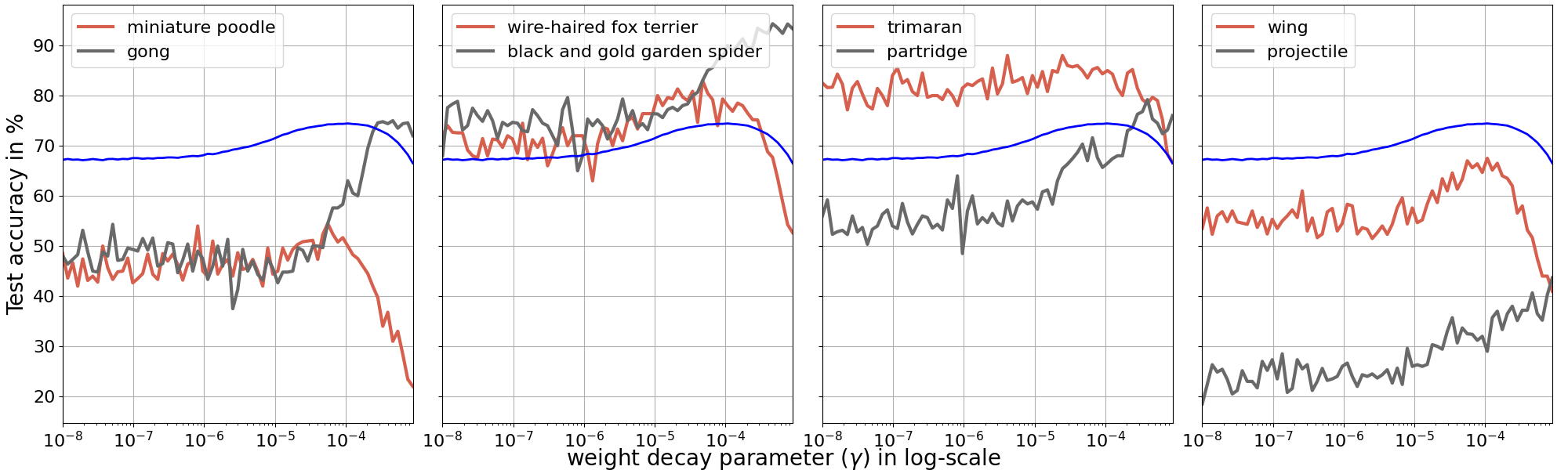}
    \vspace{-0.7cm}
    \caption{\small \textit{All results in this figure are \underline{ averaged over 10 runs} and employ the official \underline{resnet50} implementation from PyTorch that we trained on Imagenet with varying \underline{weight decay} parameter.} A very surprising result that we report here is that \textbf{the class-dependent bias that we observed from DA also occurs with one of the most fundamental and uninformed regularizer: weight decay}. We report the per-class performance when only employed horizontal flip as DA and a varying weight decay parameter, and we observe clear distinct behaviors between different classes. Samples for each class are provided in \cref{fig:resnet50_wd_samples}, in the appendix. We provide the same figure for DenseNet121 model in \cref{fig:tikhonov_densenet}.  }
    \label{fig:tikhonov}
\end{figure}

We quantified in \cref{sec:quantify} how much per-class bias was produced by DA. As per the arguments given in \cref{sec:theory,sec:pretrained}, it would be natural to assume that what makes DA responsible for creating class-dependent bias in DNs is our misfortune in defining correct augmentation policies, and thus, that other regularization techniques that are uninformed e.g. weight decay would behave differently. The goal of this section is to demonstrate that weight decay also suffers from the same class-dependent behavior indicating that designing a regularizer for DNs that is fair across classes might require novel innovative solutions.

The basic formulation of weight-decay consists in having any loss to be minimized $\mathcal{L}$ and add to it an additional term denoted as $\gamma \|\theta\|_2^2$ where $\theta$ collects the model's parameters and $\gamma$ is the strength of the regularization (proportional to the restriction on the trained model complexity). In general, we do not incorporate the bias term(s) within $\theta$ as this would directly remove the ability of the model to learn the natural class prior \citep{hastie2009elements}. As a result, and throughout this study, we consider $\theta$ to incorporate all the DN parameters except for the ones of batch-normalization layers, as commonly done in deep learning \citep{leclerc2022ffcv}. We report in \cref{fig:tikhonov} the per-class performance of a resnet50 trained on Imagenet with varying weight decay coefficient $\gamma$ (as was done for DA in \cref{fig:bias}) and we observe that different classes have different test accuracy sensitivities to variations in $\gamma$. Some will see their generalization performance increase, while others will have decreasing generalization performances. That is, {\bf even for uninformative regularizers such as weight decay a per-class bias is introduced, reducing performances for some of the classes}. The next \cref{sec:transfer} proposes to quantifying how much bias transfers to different downstream tasks.

\vspace{-0.1cm}
\subsection{The Class-Dependent Bias Transfers to Other Downstream Tasks}
\label{sec:transfer}
\vspace{-0.1cm}

The last experiment we propose is to quantify the amount of class-dependent bias that transfers to other downstream tasks, a common situation in transfer learning and in system deployment to the real world \citep{pan2009survey}. We thus want to measure how regularization applied during the pre-training phase on a {\em source} dataset impacts the per-class accuracy of that model on the {\em target} dataset.

\begin{figure}[t!]
    \centering
    \includegraphics[width=\linewidth]{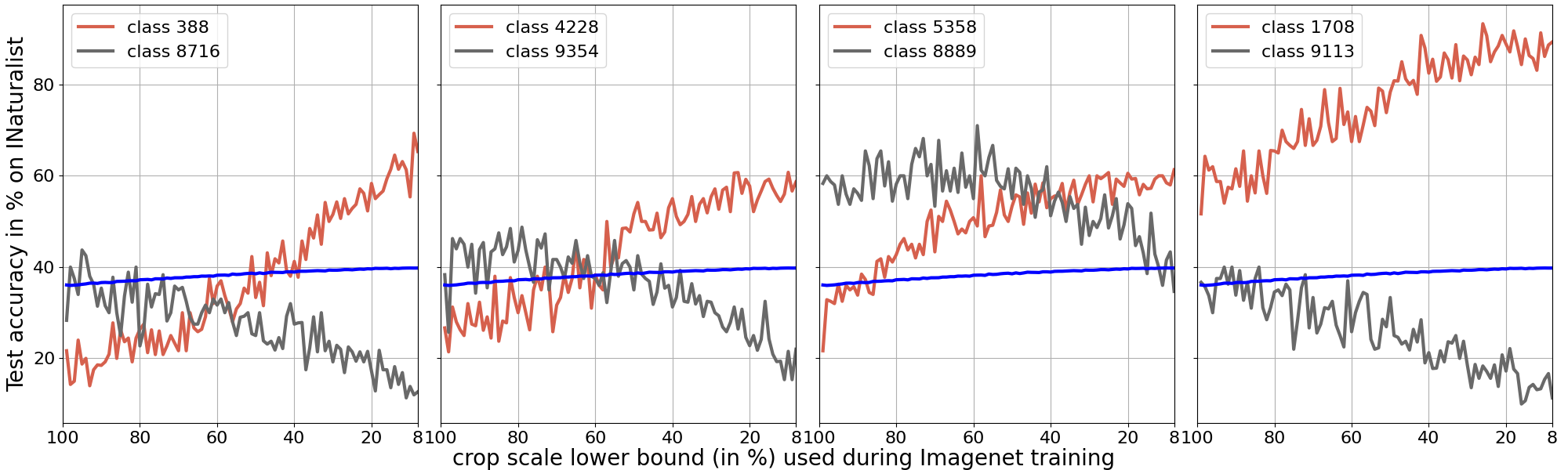}
    \vspace{-0.7cm}
    \caption{\small \textit{All results in this figure are \underline{ averaged over 10 runs} and employ the official \underline{resnet50} implementation from PyTorch that we pre-trained on Imagenet with varying \underline{random crop} lower bound and then transferred to INaturalist with frozen backbone weights (only the linear classifier is tuned).} We observe that even when the DA is applied on a different dataset during a pre-training phase, the trained model incorporates an inherent bias that transfers to downstream tasks i.e. \textbf{when deploying a model in a transfer learning task, selecting the one with best average test accuracy on the source dataset might result in deploying a model with the most biased against the classes of interest in the target dataset}. Samples for each class are provided in \cref{fig:resnet50_transfer_samples}, in the appendix.}
    \label{fig:transfer}
\end{figure}

In order to keep the setting similar to \cref{sec:quantify}, we adopt a resnet50 model with random crop DA. That model is pre-trained on Imagenet dataset (source) with varying value of $\tau$ (random crop lower bound) and then, the trained model is transferred to the INaturalist dataset \citep{van2018inaturalist} (target) that consists of 10,000 classes. When transferring the model to INaturalist, the parameters are kept frozen, and only a linear classifier is trained on top of it. We report in \cref{fig:transfer} the performance of the trained models with varying $\tau$ on different INaturalist classes. We observe once again that the best resnet50 ---on average--- is not necessarily the one that should be deployed as there exists a strong per-class bias that varies with $\tau$. As a result, picking the best performing model from a source dataset to a target dataset, might leave the pipeline to perform poorly since that model might also be the one that is the most biased against the class of interest in the target dataset.

This result should motivate the design of novel regularizers that do not reduce performances between classes at different regimes. Additionally, due to the cost of training multiple models with varying regularization settings, one might wonder on the possible alternative solutions to detect trends such as shown in \cref{fig:transfer} only when given a single pre-trained model.

\vspace{-0.1cm}
\section{Conclusion}
\vspace{-0.1cm}

We proposed in this study to understand the impact of regularization, in particular data-augmentation and weight decay, into the final performances of a deep network. We obtained that the use of regularization increases the average test performances at the cost of significant performance drops on some specific classes. By focusing on maximizing aggregate performance statistics we have produced learning mechanisms that can be potentially harmful, especially in transfer learning tasks. In fact, we have also observed that varying the amount of regularization employed during pre-training of a specific dataset impacts the per-class performances of that pre-trained model on different downstream tasks e.g. going from Imagenet to INaturalist.

\bibliographystyle{apalike}
\bibliography{refs}

\appendix

\input{appendix}

\end{document}

%% file: preamble.tex
\usepackage{amsmath,amsfonts,amsthm,amssymb,bm}

\def\eqref#1{equation~\ref{#1}}

\def\1{\bm{1}}

\def\vu{{\bm{u}}}
\def\vv{{\bm{v}}}
\def\vw{{\bm{w}}}
\def\vx{{\bm{x}}}
\def\vy{{\bm{y}}}

\def\mD{{\bm{D}}}

\def\mX{{\bm{X}}}

\DeclareMathAlphabet{\mathsfit}{\encodingdefault}{\sfdefault}{m}{sl}
\SetMathAlphabet{\mathsfit}{bold}{\encodingdefault}{\sfdefault}{bx}{n}

\def\sA{{\mathbb{A}}}

\def\sD{{\mathbb{D}}}

\def\sX{{\mathbb{X}}}

\newcommand{\Var}{\mathrm{Var}}



%% file: tikz_figure.tex
\begin{figure}[t!]
    \centering
    \begin{minipage}{0.49\linewidth}
\centering
    \underline{Current Deep Learning}\\
\begin{scaletikzpicturetowidth}{\textwidth}
\begin{tikzpicture}[scale=\tikzscale]
\tikzset{
    show curve controls/.style={
        decoration={
            show path construction,
            curveto code={
                \draw [blue, dashed]
                    (\tikzinputsegmentfirst) -- (\tikzinputsegmentsupporta)
                    node [at end, cross out, draw, solid, red, inner sep=2pt]{};
                \draw [blue, dashed]
                    (\tikzinputsegmentsupportb) -- (\tikzinputsegmentlast)
                    node [at start, cross out, draw, solid, red, inner sep=2pt]{};
            }
        }, decorate
    }
}
\tkzInit[xmax=6,ymax=3,xmin=0,ymin=0]
   \tkzDrawX[label=]\tkzDrawY[label=]
   \node[rotate=90] at (-0.3,2.7) {test error};
   \node at (5.3,-0.4) {model complexity};
   \draw [red,line width=0.45mm] plot [smooth, tension=0.6] coordinates { (0,2.5) (2.5,0.2) (6,0.9)};
    \draw [blue,line width=0.45mm] plot [smooth, tension=0.5] coordinates { (0,2.5) (1,0.2) (2,1.2) (6,1.7)};
    \draw [black!30!green,line width=0.45mm] plot [smooth, tension=0.5] coordinates { (0,2.5) (2.,0.2) (6,0.7)};
    \draw [black, dashed,line width=0.45mm] (2.45,0) -- (2.45,2.9);
    \draw [blue, dashed,line width=0.45mm] (1.05,0) -- (1.05,2.5);
    \draw [red, dashed,line width=0.45mm] (2.9,0) -- (2.9,2.2);
    \draw [black!30!green, dashed,line width=0.45mm] (2.2,0) -- (2.2,1.9);
    
\node[font={\small\bfseries}] at (2.45,3.2) {best on average};
\node[color=red,font={\small\bfseries}] at (3.1,2.4) {best 1};
\node[color=blue,font={\small\bfseries}] at (1.02,2.65) {best 0};
\node[color=black!30!green,font={\small\bfseries}] at (1.8,2.1) {best 2};
\node[color=red,font={\small\bfseries}] at (5.5,2.5) {class 1};
\node[color=blue,font={\small\bfseries}] at (5.5,2.8) {class 0};
\node[color=black!30!green,font={\small\bfseries}] at (5.5,2.2) {class 2};

\path[draw=black, line width=0.4mm, -{Triangle[length=1mm, bend]}, shorten >=1mm, shorten <=0.5mm]    
        (1.5,-0.3) to [bend left]    (2.5,0.5);
\node[text width=4cm,font={\small\bfseries},align=center] at (1.5,-0.8) {best average test error of $\delta$ with poor generalization on \color{blue}{class 0}};
\end{tikzpicture}
\end{scaletikzpicturetowidth}
\end{minipage}
\begin{minipage}{0.49\linewidth}
\centering
    \underline{Ideal Complexity Measure}\\
\begin{scaletikzpicturetowidth}{\textwidth}
\begin{tikzpicture}[scale=\tikzscale]
\tikzset{
    show curve controls/.style={
        decoration={
            show path construction,
            curveto code={
                \draw [blue, dashed]
                    (\tikzinputsegmentfirst) -- (\tikzinputsegmentsupporta)
                    node [at end, cross out, draw, solid, red, inner sep=2pt]{};
                \draw [blue, dashed]
                    (\tikzinputsegmentsupportb) -- (\tikzinputsegmentlast)
                    node [at start, cross out, draw, solid, red, inner sep=2pt]{};
            }
        }, decorate
    }
}
\tkzInit[xmax=6,ymax=3,xmin=0,ymin=0]
   \tkzDrawX[label=]\tkzDrawY[label=]
   \node[rotate=90] at (-0.3,2.7) {test error};
   \node at (5.3,-0.4) {model complexity};
   \draw [red,line width=0.45mm] plot [smooth, tension=0.6] coordinates { (0,2.5) (2.5,0.2) (6,0.9)};
    \draw [blue,line width=0.45mm] plot [smooth, tension=0.5] coordinates { (0,2.5) (2.3,0.2) (4,0.7) (6,1.7)};
    \draw [black, dashed,line width=0.45mm] (2.75,0) -- (2.75,2.4);
    \draw [black!30!green,line width=0.45mm] plot [smooth, tension=0.5] coordinates { (0,2.5) (2.,0.2) (6,0.7)};

\path[draw=black, line width=0.4mm, -{Triangle[length=1mm, bend]}, shorten >=1mm, shorten <=0.5mm]    
        (1.5,-0.3) to [bend left]    (2.8,0.2);
\node[text width=4.2cm,font={\small\bfseries},align=center] at (1.5,-0.8) {best average test error of $\gamma \ll \delta$ with similar inter-class generalization};
\node[font={\small\bfseries},text width=4.2cm,align=center] at (2.75,2.8) {agreement on the optimal complexity};

\node[color=red,font={\small\bfseries}] at (5.5,2.5) {class 1};
\node[color=blue,font={\small\bfseries}] at (5.5,2.8) {class 0};
\node[color=black!30!green,font={\small\bfseries}] at (5.5,2.2) {class 2};
\end{tikzpicture}
\end{scaletikzpicturetowidth}
\end{minipage}
    \caption{\small Structural risk minimization minimizes the empirical risk of several models with varying complexity, and selects the one offering the best compromise between under-fitting and over-fitting \citep{vapnik1974method}. In deep learning, controlling the model's complexity is achieved by picking different DN architectures and/or by applying different levels and flavors of regularization. The key observation of our study is that {\bf when the model complexity is calibrated by a regularizer such as DA (see \cref{fig:dataset_accuracy,fig:bias,fig:allmodels}) or weight-decay (see \cref{fig:tikhonov}), the class-conditional empirical risks do not align between classes i.e. cross-validation produces models that perform really well on the majority of classes but arbitrarily poorly on the others} as depicted on the left-hand-side. In an ideal setting where the control of the model's complexity is well aligned with the task and  model, one would observe the right-hand-side scenario where the same model complexity is optimal for all classes. 
    }
    \label{fig:teaser}
\end{figure}

%% file: appendix.tex
\newpage

\begin{center}
\rule{12cm}{0.1cm}\\[1em]
{\Huge Supplementary Materials}\\[1em]
\rule{12cm}{0.1cm}
\end{center}

\section{Theoretical Relations Between Data-Augmentation, Generalization, and a Model's Bias and Robustness}
\label{sec:more_theory}

\subsection{Data-Augmentation, Regularization and Structural Risk Minimization}

\textbf{Bias-Variance Decomposition.}~
The expected error of a trained model can {\em always} be decomposed into two terms that can be tune by altering the considered model class and regularization, and a third term which is the inherent measurement noise, that can not be reduced. In fact, the true labels are obtained from $\vy=f(\vx)+\epsilon$ with $f$ the true, unknown, data model, and $\epsilon$ some irreducible error i.e. coming from measurements or data compression. For the Mean-Squared Error (MSE) we obtain this decomposition as
\begin{align*}
    \mathbb{E}_{\epsilon,\mX,\sD}\left[\left(\vy - \hat{f}(\vx;\mD)\right)^2\right]=&\mathbb{E}_{\epsilon,\mX,\sD}\left[\epsilon^2+ \epsilon\left(f(\vx) - \hat{f}(\vx;\mD)\right)+ \left(f(\vx) - \hat{f}(\vx;\mD)\right)^2\right]\\
    =&\Var(\epsilon)^2+\mathbb{E}_{\mX}\left[\left(f(\vx) - \mathbb{E}_{\sD}\left[\hat{f}(\vx;\mD)\right]\right)^2-\mathbb{E}_{\sD}\left[\hat{f}(\vx;\mD)\right]^2+\mathbb{E}_{\sD}\left[\hat{f}(\vx;\mD)^2\right]\right]\\
    =&\Var(\epsilon)^2+\mathbb{E}_{\mX}\bigg[\underbrace{\left(f(\vx) - \mathbb{E}_{\sD}\left[\hat{f}(\vx;\mD)\right]\right)^2}_{\text{bias}}+\underbrace{\Var_{\sD}\left(\hat{f}(\vx;\mD)\right)}_{\text{variance}}\bigg].
\end{align*}
Although we only derive here the MSE case, the same can be obtained in the multivariate case and in classification settings \citep{valentini2004bias,hastie2009elements}. When the model is with low complexity e.g. the model is under-parametrized, or the regularization is applied aggressively, the variance term becomes small and the bias term increases. Conversely, when the model is not regularized and is overparametrized, the variance term increases and the bias term reduces. One strategy to find the best model is offered by structural risk minimization.

\textbf{Structural Risk Minimization (SRM).}~
As introduced by \citet{vapnik1974method}, SRM proposes a search strategy to obtain the best model. One first chooses a class of function that $\hat{f}$ must live in, hopefully informed from a priori knowledge e.g. polynomials of degree $k$, resnet50 DN architecture. Then, one finds a hierarchical construction of nested functional spaces that relate to the function complexity. For example, it could be considering polynomials of increasing degree, up to $k$, or considering resnet50s with weights being bounded by an increasing constant. Finally, one minimizes the empirical risk (training error) for each model and picks the one with best valid set performances. The valid set performance can be estimated empirically from a subset of the training set that has been set apart before fitting, or from other measures such as the VC-dimension of the trained model \citep{vapnik2015uniform}. For example, cross-validating the weight decay parameter of a model corresponds to SRM. We make this connection precise below.

\textbf{From Data-Augmentation and Weight Decay to Nested Functional Spaces.}Regularization is commonly employed during training to prevent overfitting i.e. reduce the model complexity. One might argue that regularization does not {\em strictly} restrict the model functional space, it simply {\em favors} simpler models to be used. Yet, it turns out that regularization can be cast as explicitly restricting the parameter space of the model i.e. restricting the functional space of the model.

To see this, let's consider the following loss function to be minimized $\ell(\theta,\gamma)$ that depends of the parameters $\theta$ and the dataset $\mD$. We define the following two optimization problems, one with Tikhonov regularization with amplitude $\gamma$ and one with a constraint on the space of $\theta$
\begin{align*}
    \min_{\theta \in \mathbb{R}^{K}} \ell(\theta,\mD) + \gamma \|\theta \|_2^2,&&(P1)\\
    \min_{\theta \in \mathbb{R}^{K}:\|\theta\|_2^2\leq\beta} \ell(\theta,\mD).&&(P2)
\end{align*}
Let's also denote by $\hat{\theta}(\gamma)$ the parameter value that is a global minimum of (P1); if multiple global minimum exist, pick the one with greatest $\ell_2$ norm. We now have to solve for $\theta,\lambda$ the following system that results from the Lagrangian, with $\lambda$ the Lagrange multiplier \citep{bertsekas2014constrained} and $s^2$ the slack variable of the inequality
\begin{align*}
    \nabla_{\theta}\ell(\theta,\mD)+ 2\lambda\theta=0 \text{ and }
    \|\theta\|_2^2-\beta-s^2=0\text{ and }2\lambda s=0,
\end{align*}
setting $\lambda \leftarrow \gamma$, $\theta \leftarrow \hat{\theta}(\gamma)$, and $\beta \leftarrow \|\hat{\theta}(\gamma) \|_2^2,s\leftarrow 0$ solves the system. As a result, solving the constrained optimization problem with $\beta = \|\hat{\theta}(\gamma)\|_2^2$ and solving the Tikhonov regularized problem with $\gamma$ coefficient is equivalent. Since we also have that $\|\hat{\theta}(\gamma_1)\|_2^2\leq \|\hat{\theta}(\gamma_2)\|_2^2$, we obtain that starting from a high value of Tikhonov regularization and gradually reducing it produces nested functional spaces where the model live in. More general results can be obtained in different regularization settings in \citet{NIPS2009_d516b136,hastie2009elements}. Moving to the case of DA, the same procedure applies. In fact, one can use the results of \citet{phaisangittisagul2016analysis,lejeune2019implicit,balestriero2022data} to cast DAs such as dropout, and image perturbations as explicit regularizers \`a la Tikhonov and repeat the above procedure. In fact, DA can be seen as producing a model with lower variance and a bias towards the augmented dataset. Hence, {\bf if the augmented dataset is not aligned with the true underlying data distribution, DA will increase the model's bias}. For results studying DA in the context of structural risk minimization, we refer the reader to \citet{chapelle2000vicinal,arjovsky2019invariant,chen2020group,ilse2021selecting}. In short, it is clear from the above that DA is yet another form of regularization that can be applied for structural risk minimization. Yet, even if DA effectively restrict the functional space of the model, it can be at the cost of producing strong biases, as we empirically observed in \cref{sec:DA}.

{\bf Provable Bias Induced by Data-Augmentation.} Especially relevant to our results is a recent result of 
\citet{xu2020wemix}. In this work, it was theorized what when the underlying dataset contains inherent biases, training on the original data is more effective than employing an i.i.d. DA policy, i.e. applying the same random augmentation to all samples/classes, to produce an unbiased model. In short, the DA policy exacerbates the already present biases and makes the trained model further away from the unbiased optimum. As per our experiments from \cref{sec:DA}, we observe that bias can take many form. For example, having classes with different statistics e.g. objects that appear at very specific scale in the image. One strategy proposed by \citet{xu2020wemix}, assuming that the bias of a model can be measured, was to adapt the DA policy to the dataset at hand to actively correct the present bias, as done for example in \citet{mclaughlin2015data,iosifidis2018dealing,jaipuria2020deflating}. In fact, when the bias is measurable, many recent studies have shown that DAs could be designed to reduce the bias present in a dataset \citep{mclaughlin2015data,jaipuria2020deflating}.

\subsection{Known Tradeoff Between Regularization, Generalization and Robustness}

There exist a few previous work in that direction, especially in the field of robust and overparametrized machine learning.

In \citet{raghunathan2020understanding} a surprising result demonstrated that the minimum norm interpolant of the original + DA could have a larger standard error than that of the original data's minimum norm interpolant. Even more surprising, this result was obtained using  consistent DA i.e. transformations that do not alter the label information of the samples as in $p_{\vy|\vx}=p_{\vy|\mathcal{T}_{\alpha}(\vx)}$, or $f^*(\vx)=f^*(\mathcal{T}_{\alpha}(\vx))$ as discussed in \cref{eq:ratio}. This possible degradation of performance occurs as long as the model remains over-parametrized, even when considering the original+DA dataset.

In addition to the implication of DA into bias, there exists an intertwined relationship between DA and model robustness. In fact, even assuming the use of perfectly adapted DAs, there exists an {\em inherent tradeoff} between accuracy and robustness that holds even in the infinite data limit
\citep{tsipras2018robustness,fawzi2018analysis,zhang2019theoretically}. For example, \citet{min2021curious} proved in the robust linear classification regime that (i) more data improves generalization in a weak adversary regime, (ii) more data can improve generalization up to a point where additional data starts to hurt generalization in a medium adversary regime, and that (iii) more data immediately decreases generalization error in a strong adversary regime.

Lastly, a few studies have started to study the bias the could be cause from some DAs, although those studies focused on learned DAs e.g. from Generative Adversarial Networks \citep{hu2019exploring}. In that scenario, the predicament is that the GAN in itself is biased, and thus any GAN-generated DA will inherit those biases.

\section{Proof of \cref{thm:bias}}
\label{sec:proof_bias}
\begin{proof}
To streamline our derivations and without loss of generality, we first impose that $\mathcal{T}_{-\alpha}$ inverts the action of $\mathcal{T}_{\alpha}$, that $\mathcal{T}_0$ acts as the identity mapping and that $\mathcal{T}_{\alpha} \circ \mathcal{T}_{\beta}=\mathcal{T}_{\alpha+\beta}$. In short, we have a {\em group} structure that allows us to define the following equivalence class $\sim$ that is positive iff two images are related by a transformation, formally defined as
\begin{align}
    \vu \sim \vv \iff \exists \alpha : T_{\alpha}(\vu)=\vv,\label{eq:equivalence}
\end{align}
which is {\em reflexive} ($\vu \sim \vu$ with $\alpha=0$), {\em symmetric} ($\vu \sim \vv$ with $\alpha$ implies that $\vv \sim \vu$ with $-\alpha$), and {\em transitive} ($\vu \sim \vv$ with $\alpha$ and $\vv \sim \vw$ with $\beta$ implies that $\vu \sim \vw$ with $\alpha+\beta$). 

Lastly, given a set of samples $\sX$, we define the following $a$ mapping that will decompose $\sX$ into a set generators and corresponding DA parameters that allow to recover those subsets when applied to the subset generator. Formally, we have
\begin{align}
    g(\sX) \mapsto \left\{(\mu_1,\sA_1),(\mu_2,\sA_2),\dots \right\} \text{ s.t. } \cup_{(\mu,\sA)\in g(\sX)}\{T_{\alpha}(\mu),\forall \alpha \in \sA \} = \sX.
\end{align}
Note that there is a bijection between the pairs given by $g(\sX)$ and the set of equivalent samples defined by $\sX / \sim$. Using this, we obtain the following decomposition of the empirical error
\begin{align*}
    \mathcal{L}=\mathbb{E}_{\mX}\left[\|f^{*}(\vx)-f_{\theta}(\vx)\|\right]
    =\sum_{(\mu,\sA) \in g(\sX)}\;\;\sum_{\alpha \in \sA}\|f^{*}(\mathcal{T}_{\alpha}(\mu))-f_{\theta}(\mathcal{T}_{\alpha}(\mu))\|p(\mathcal{T}_{\alpha}(\mu)).
\end{align*}
Given the above, we now simply split a dataset $\sX$ into two, one that contains training samples, and one that contains everything else. Without loss of generality we assume that $p(\mathcal{T}_{\alpha}(\mu))$ is a constant and thus omit it. Using the fact that the model has $0$ training error, we obtain that the approximation error between $f^*$ and $f_{\theta}$ can not be $0$ unless the DA is perfectly aligned with the level-sets of $f^*$ as in
\begin{align*}
    \mathcal{L}
    =&\sum_{(\mu,\sA) \in g(\sX\setminus \sX_{\rm train})}\;\;\sum_{\alpha \in \sA}\|f^{*}(\mathcal{T}_{\alpha}(\mu))-f_{\theta}(\mathcal{T}_{\alpha}(\mu))\|_2^2+\sum_{(\mu,\sA) \in g(\sX_{\rm train})}\;\;\sum_{\alpha \in \sA}\|f^{*}(\mathcal{T}_{\alpha}(\mu))-f_{\theta}(\mathcal{T}_{\alpha}(\mu))\|_2^2\\
    \geq &\sum_{(\mu,\sA) \in g(\sX_{\rm train})}\;\;\sum_{\alpha \in \sA}\|f^{*}(\mathcal{T}_{\alpha}(\mu))-f_{\theta}(\mathcal{T}_{\alpha}(\mu))\|_2^2\\
    \geq&\sum_{(\mu,\sA) \in g(\sX_{\rm train})}\;\;\sum_{\alpha \in \sA}\|f^{*}(\mathcal{T}_{\alpha}(\mu))-{\rm const}\|_2^2,
\end{align*}
where the last equality assumes that the model has $0$ training error, and will thus predict the same constant for all DA version of the same sample. And since that last equation is always positive if the DA does not respect the true level-sets of the true function $f^*$ we obtain the desired result.
The same derivation can be carried out with any desired loss e.g. for classification tasks.
\end{proof}

\section{Additional Figures}

\begin{figure}[t!]
    \centering
    \begin{minipage}{0.49\linewidth}
    \includegraphics[width=\linewidth]{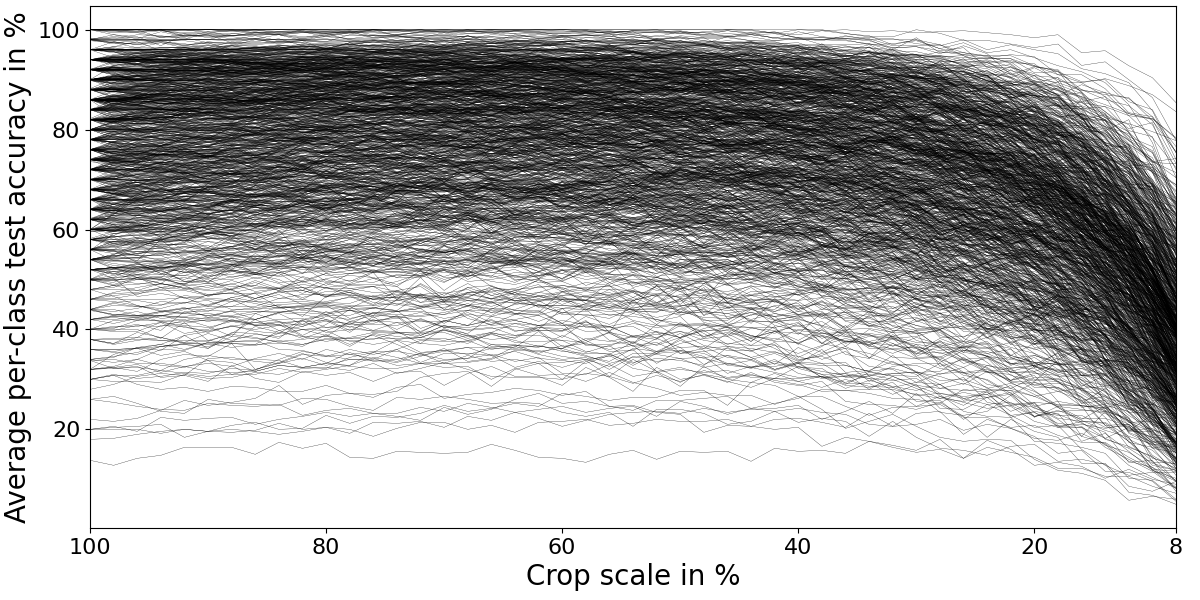}
    \end{minipage}
    \begin{minipage}{0.49\linewidth}
    \caption{\small Per-class accuracy evolution when the proportion of the image being cropped ($\theta$) is progressively increased from $8$ to $100$. While the overall trends mostly applies across classes, some extreme cases can present different sensitivity to aggressive cropping, leading to some classes being more successful as being label-preserved after application of the cropping. The weighted average of those lines (based on the proportion of samples coming from each class) gives back \cref{fig:dataset_accuracy}).
    }
    \label{fig:allclass_accuracy}
    \end{minipage}
\end{figure}


\begin{figure}[h]
    \centering
    \includegraphics[width=\linewidth]{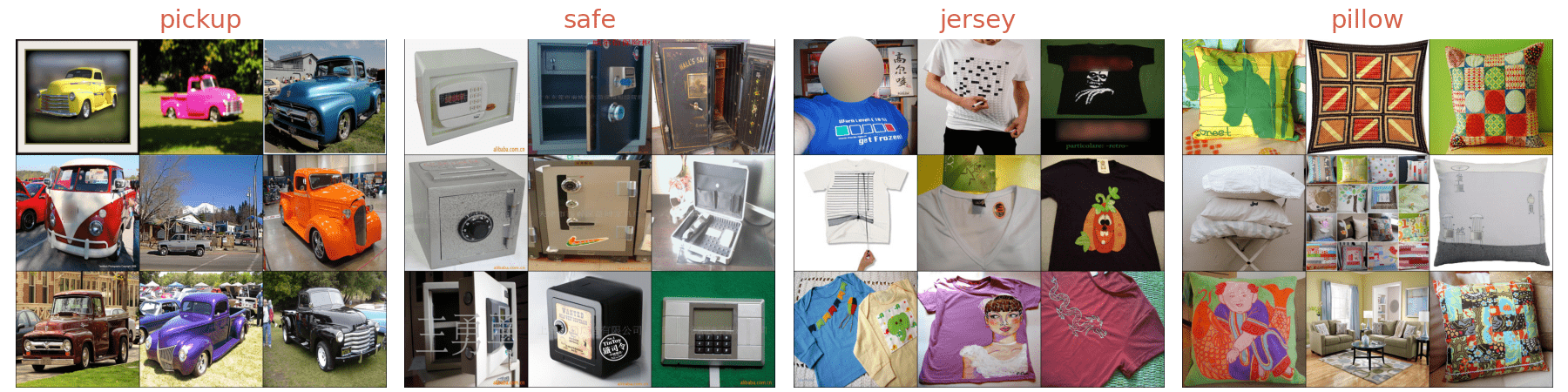}\\
    \includegraphics[width=\linewidth]{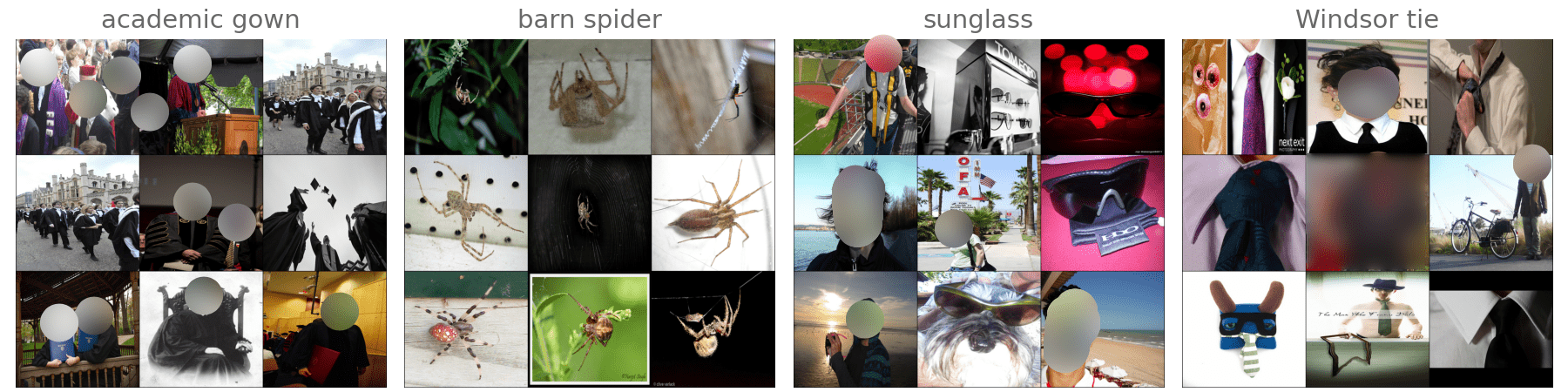}\\
    \caption{\small Random samples from the validation set of Imagenet that correspond to \cref{fig:bias}.}
    \label{fig:resnet50_crop_samples}
\end{figure}

\begin{figure}[h]
    \centering
    \includegraphics[width=\linewidth]{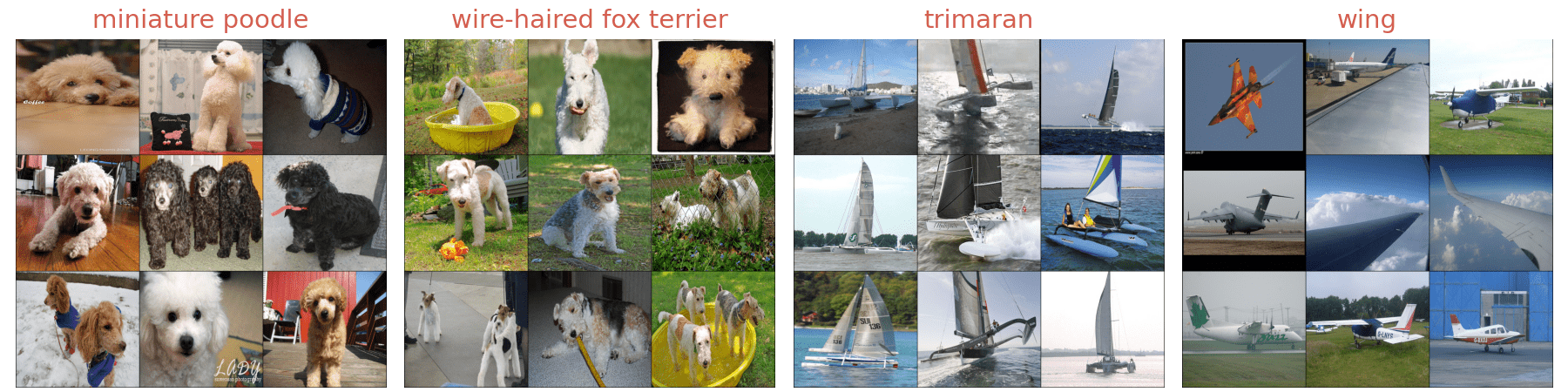}\\
    \includegraphics[width=\linewidth]{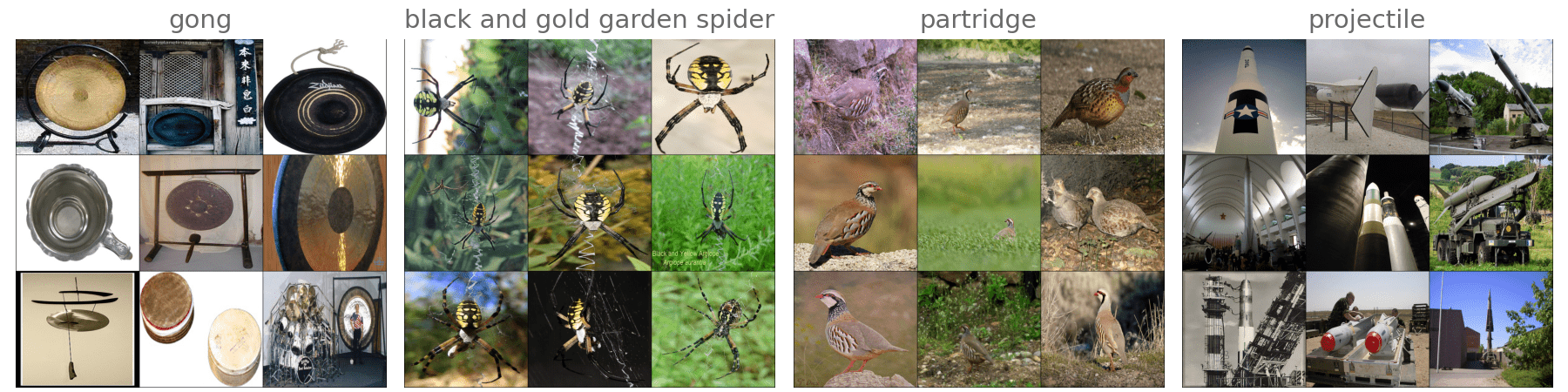}\\
    \caption{\small Random samples from the validation set of Imagenet that correspond to \cref{fig:tikhonov}.}
    \label{fig:resnet50_wd_samples}
\end{figure}

\begin{figure}[h]
    \centering
    \includegraphics[width=\linewidth]{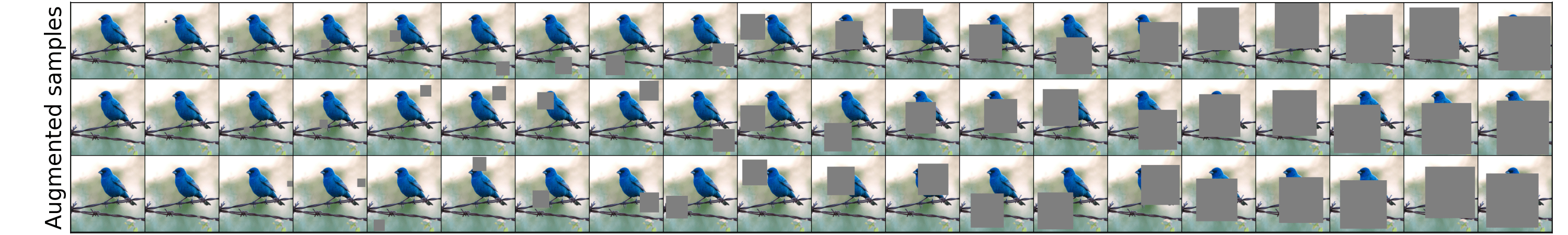}\\[0em]
    \includegraphics[width=\linewidth]{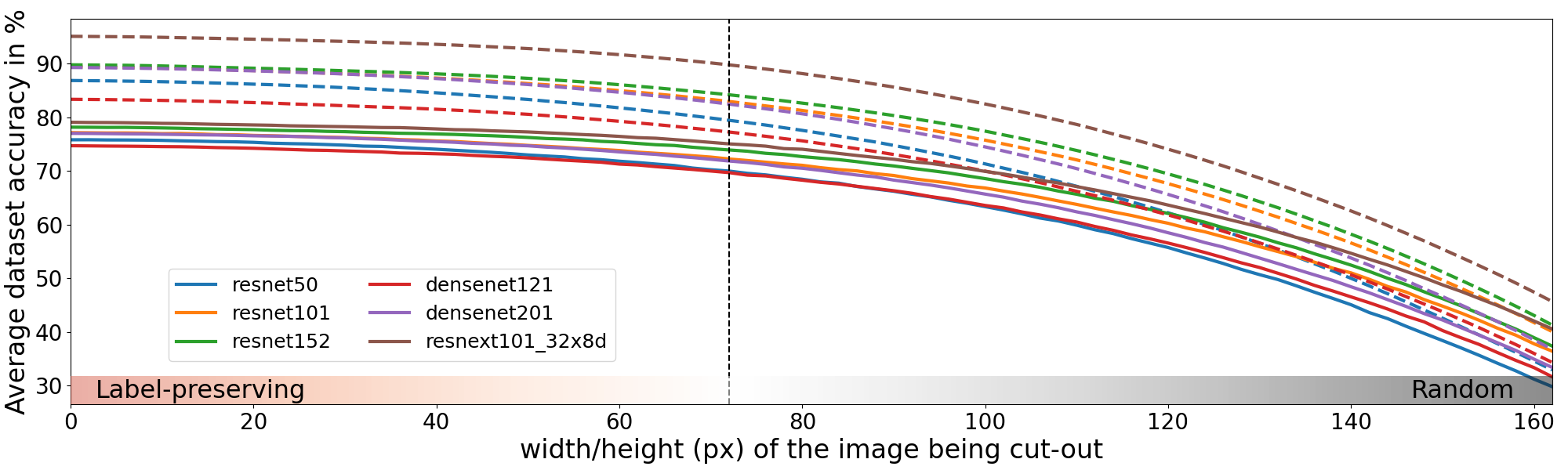}\\
    \includegraphics[width=0.66\linewidth]{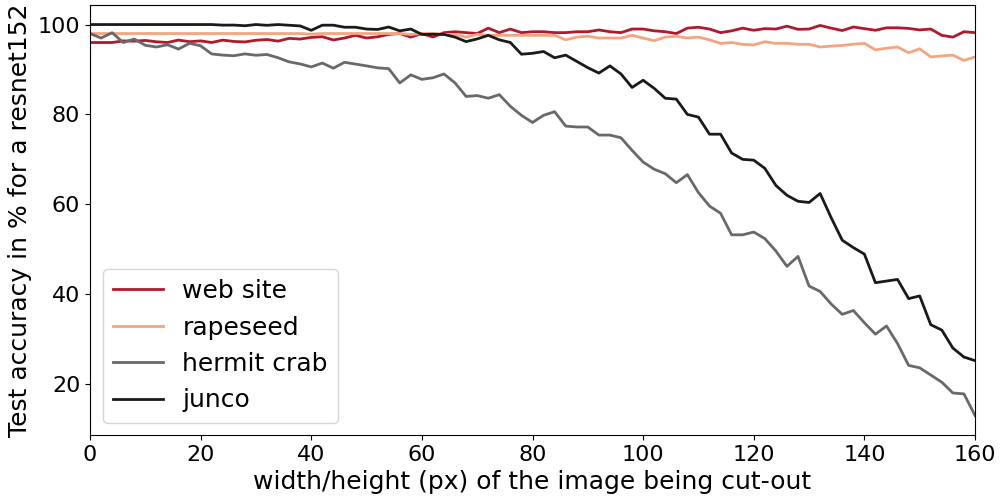}
    \includegraphics[width=0.33\linewidth]{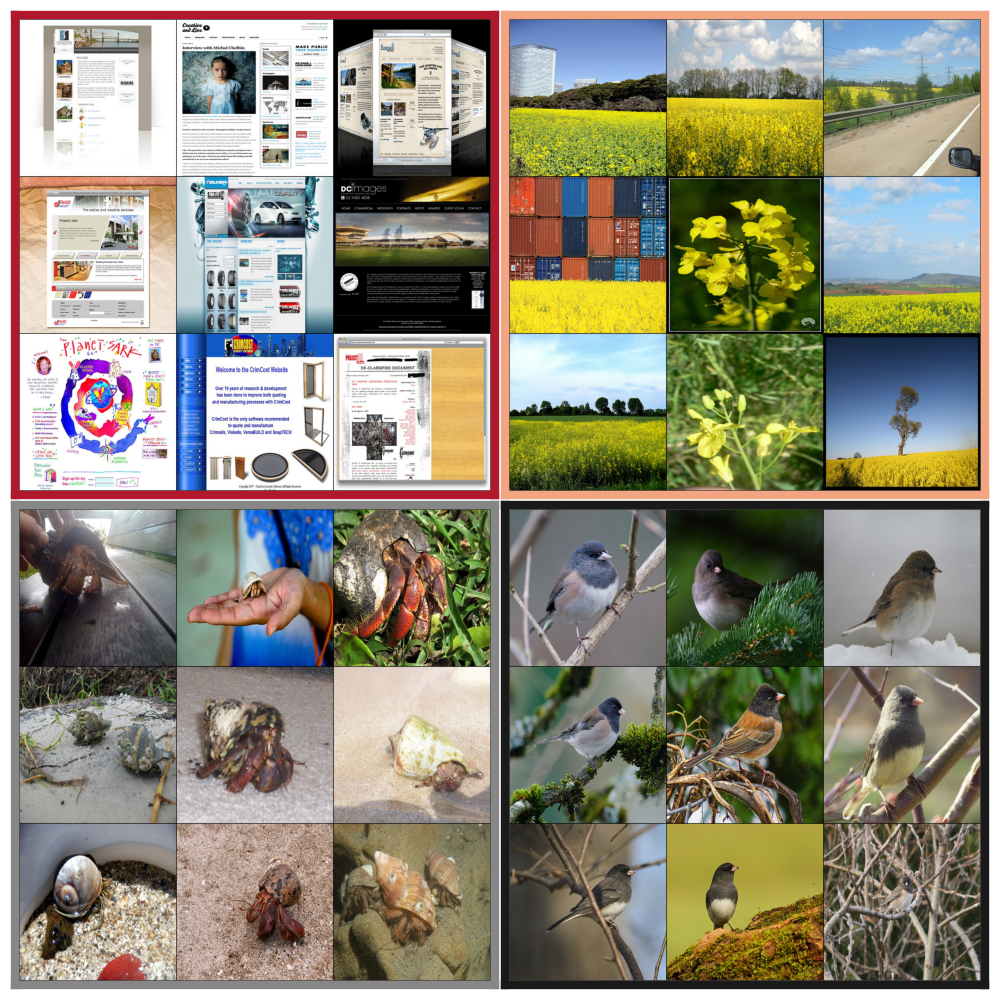}
    \vspace{-0.7cm}
    \caption{\small Reprise of \cref{fig:dataset_accuracy} but using color jittering DA. The same conclusions are reached, different classes have different label-preserving regimes under this DA.}
    \label{fig:dataset_erasing}
\end{figure}

\begin{figure}[t!]
    \centering
    \includegraphics[width=\linewidth]{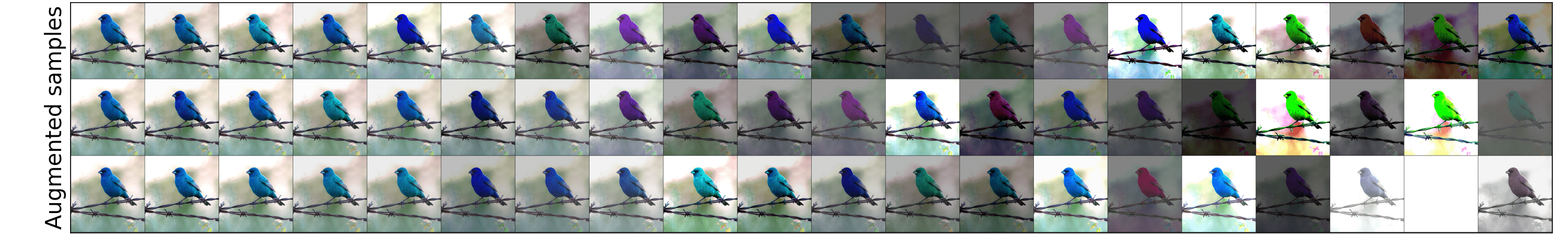}\\[0em]
    \includegraphics[width=\linewidth]{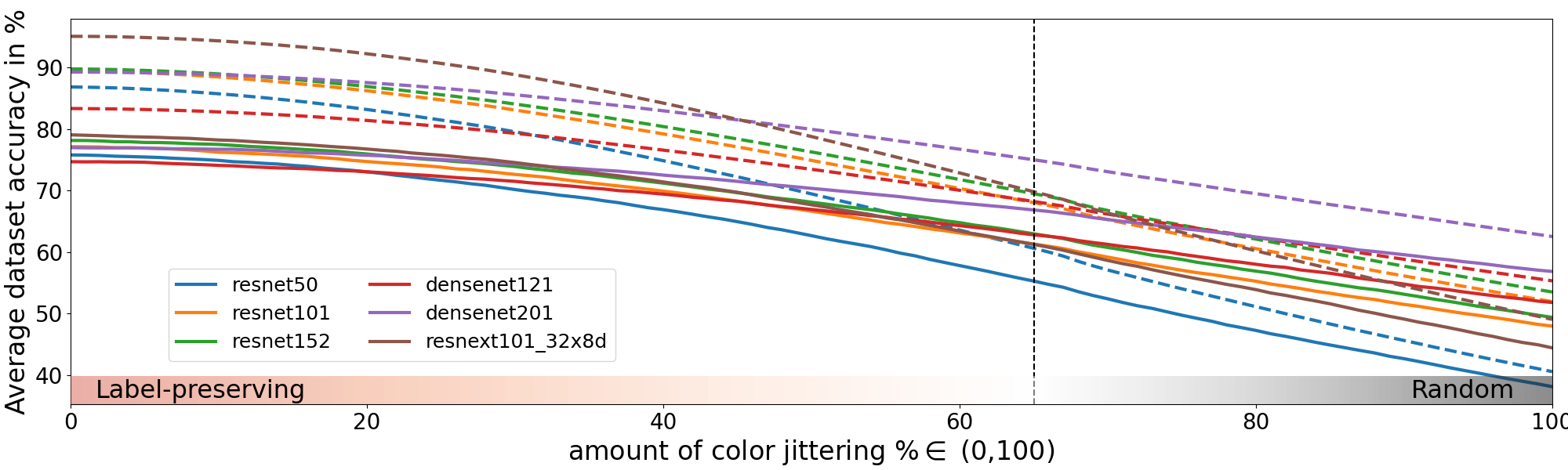}\\
    \includegraphics[width=0.66\linewidth]{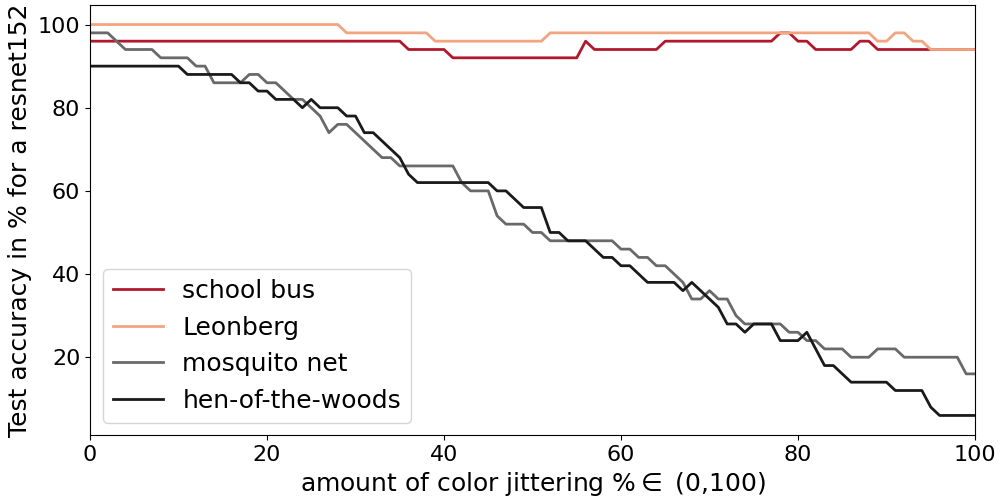}
    \includegraphics[width=0.33\linewidth]{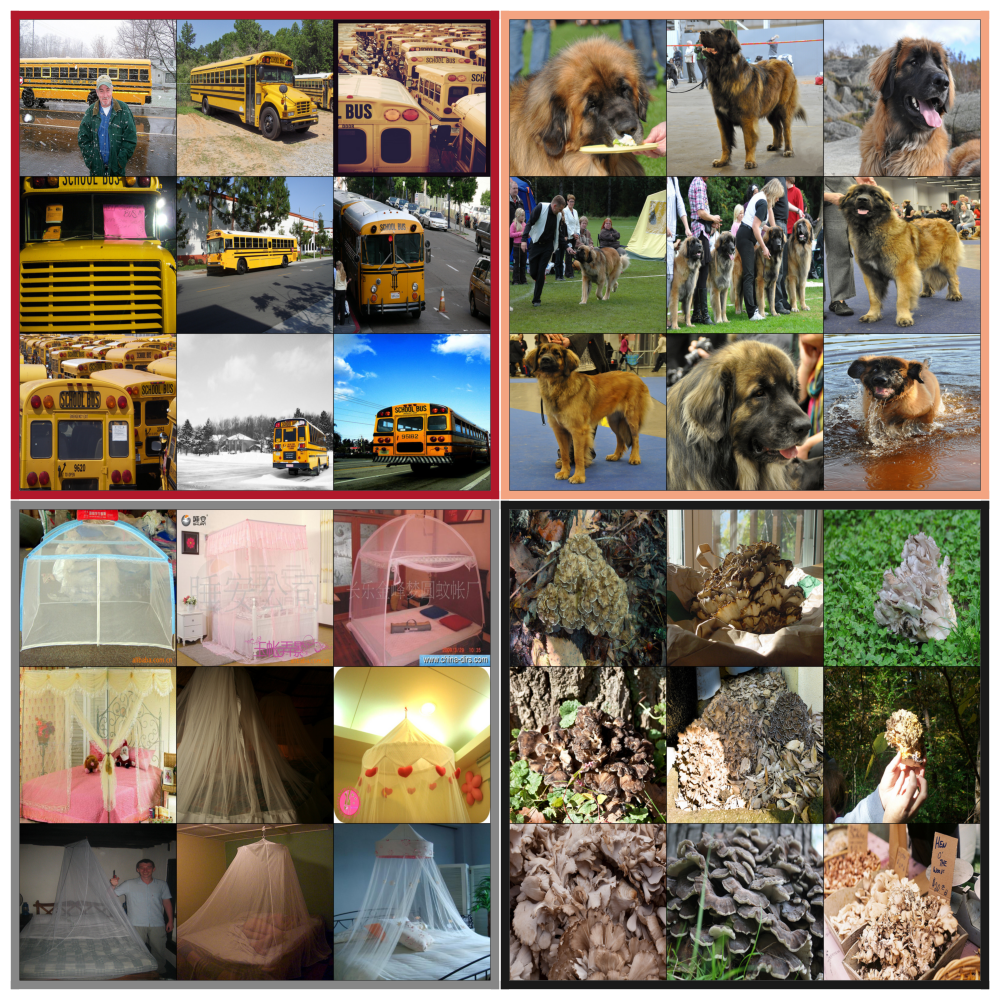}
    \vspace{-0.7cm}
    \caption{\small Reprise of \cref{fig:dataset_accuracy} but using color jittering DA. The same conclusions are reached, different classes have different label-preserving regimes under this DA.}
    \label{fig:dataset_color}
\end{figure}

\begin{figure}[ht]
    \centering
    \includegraphics[width=\linewidth]{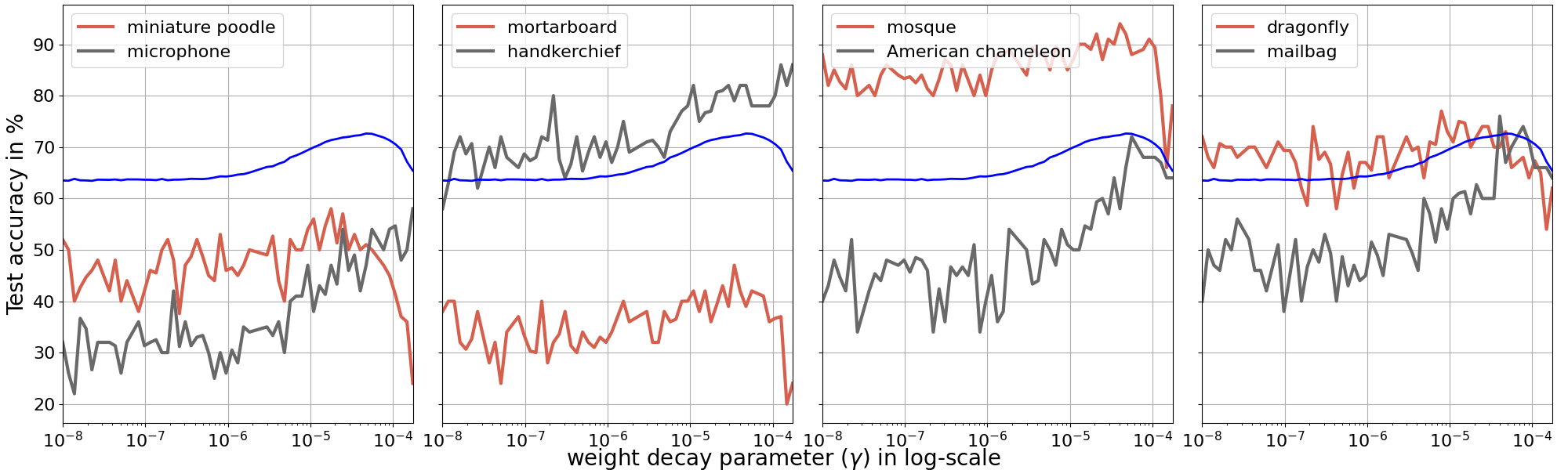}\\
    \includegraphics[width=\linewidth]{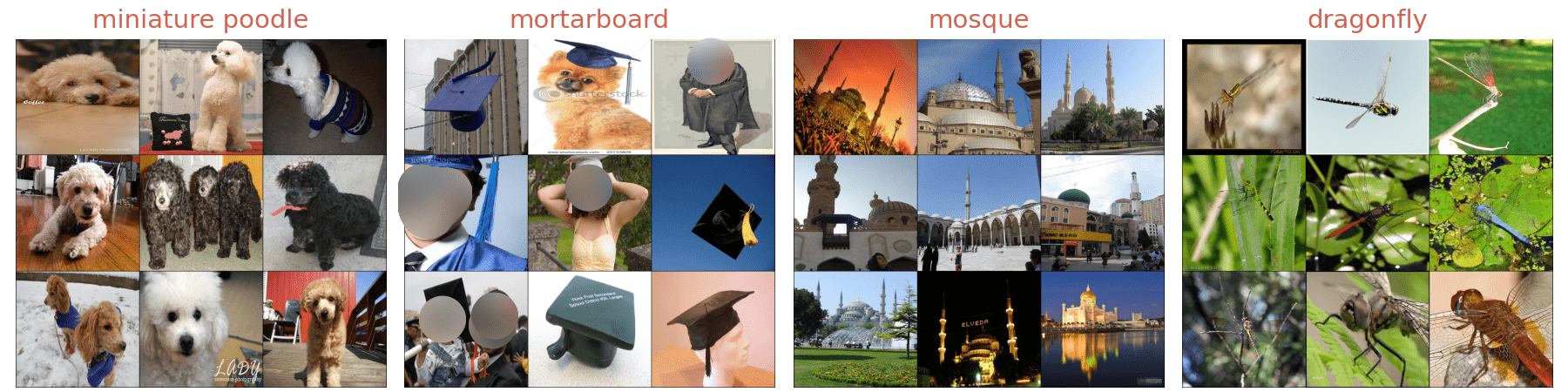}\\
    \includegraphics[width=\linewidth]{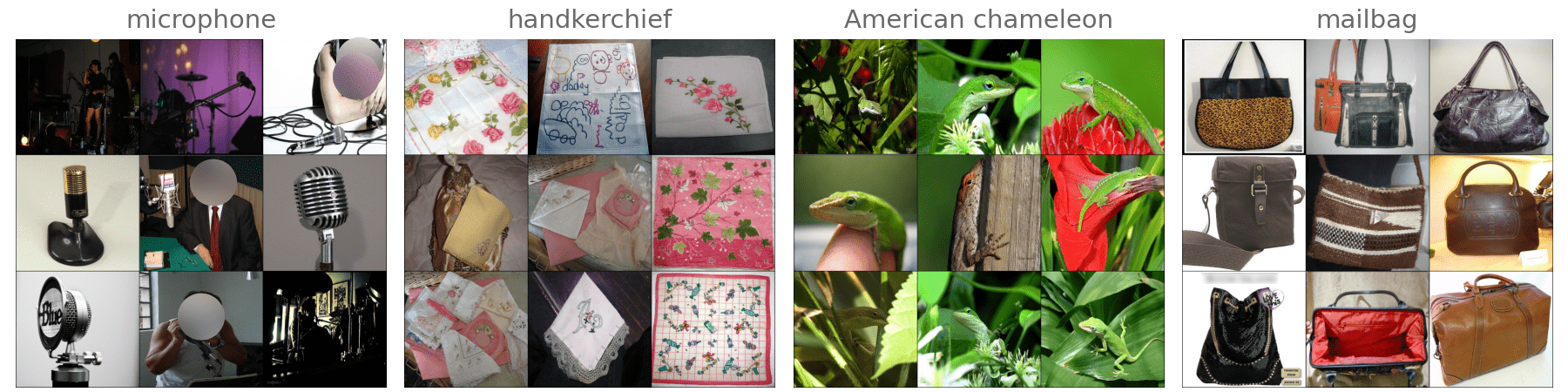}
    \caption{\small \textit{All results in this figure employ official model implementations from PyTorch that we have trained on Imagenet with varying weight decay parameter.} Reprise of \cref{fig:tikhonov} but now with a different architecture (DenseNet121) instead of the ResNet50.}
    \label{fig:tikhonov_densenet}
\end{figure}

\begin{figure}[h]
    \centering
    \includegraphics[width=\linewidth]{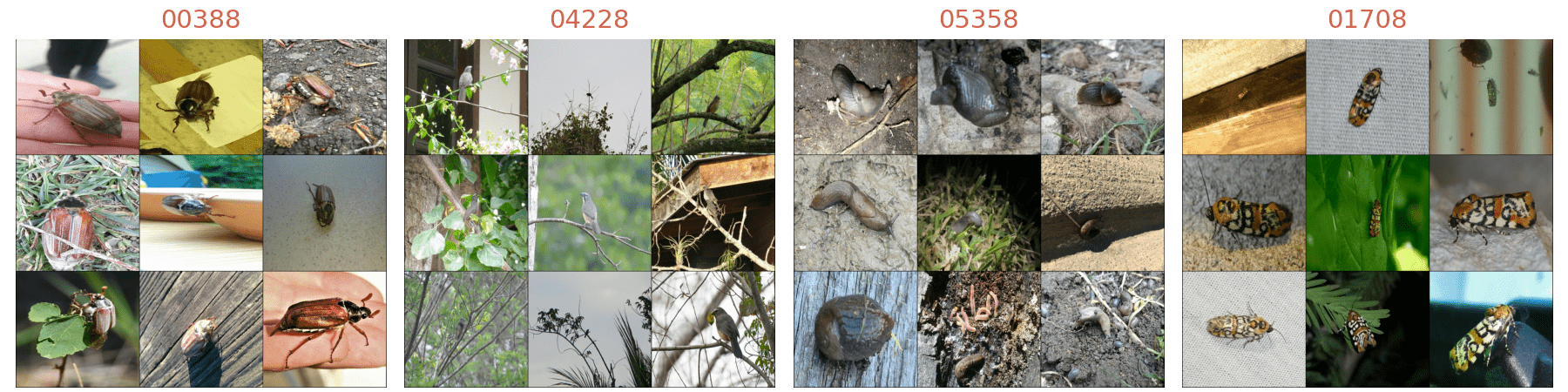}\\
    \includegraphics[width=\linewidth]{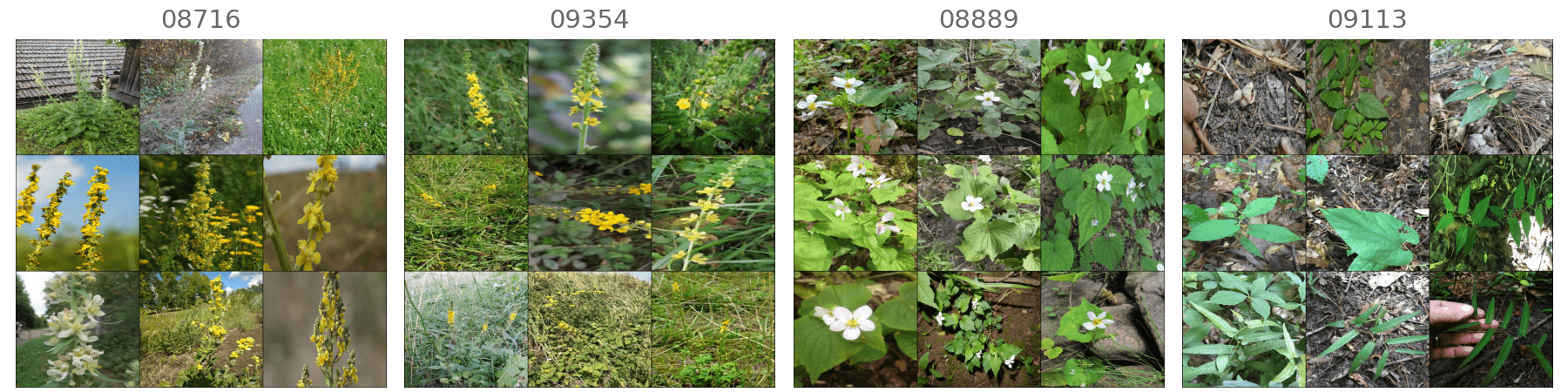}\\
    \caption{\small Random samples from the mini training set of INaturalist that correspond to \cref{fig:transfer}}
    \label{fig:resnet50_transfer_samples}
\end{figure}